\documentclass[11pt]{article}
\usepackage[final]{acl}

\usepackage{times}
\usepackage{latexsym}
\usepackage[T1]{fontenc}
\usepackage[utf8]{inputenc}
\usepackage{microtype}
\usepackage{inconsolata}
\usepackage{graphicx}
\usepackage{array}
\usepackage{booktabs}
\usepackage{amsfonts}
\usepackage{amsmath}
\usepackage{nicefrac}
\usepackage{xcolor}
\usepackage{multirow}
\usepackage{enumitem}
\usepackage{subcaption}
\usepackage{algorithm}
\usepackage{algpseudocode}
\usepackage{wrapfig}
\usepackage{float}
\usepackage{tcolorbox}
\tcbuselibrary{breakable}
\usepackage{listings}

\graphicspath{{figures/}}

\newcommand{\fa}{\textsc{FinAcumen}}
\newcommand{\fm}{\textsc{FM}}
\newcommand{\ft}{\textsc{FT}}

\lstdefinestyle{promptstyle}{
  language={},
  basicstyle=\ttfamily\footnotesize,
  breaklines=true,
  columns=fullflexible,
  showstringspaces=false,
  keepspaces=true,
  frame=none,
}

\title{FinAcumen: Financial Multimodal Reasoning via Self-Evolving Experience Memory Harness}

\author{
  \textbf{Pianran Guo\textsuperscript{1,*}},
  \textbf{Pengcheng Zhou\textsuperscript{2,*}},
  \textbf{Yuchen Jian\textsuperscript{1}},
  \textbf{Shuhua Chen\textsuperscript{1}},
\\
  \textbf{Zhongliang Yang\textsuperscript{1,$\dagger$}},
  \textbf{Linna Zhou\textsuperscript{1,$\dagger$}}
\\
\\
  \textsuperscript{1}Beijing University of Posts and Telecommunications
\\
  \textsuperscript{2}Queen Mary University of London
\\
  \texttt{guopianran@bupt.edu.cn},
  \texttt{jp2020213547@qmul.ac.uk}
\\
  \texttt{jianyucheng@bupt.edu.cn},
  \texttt{lyonia\_orangerica@bupt.edu.cn}
\\
  \texttt{yangzl@bupt.edu.cn},
  \texttt{zhoulinna@bupt.edu.cn}
\\
  \textsuperscript{$\dagger$}Corresponding authors: yangzl@bupt.edu.cn, zhoulinna@bupt.edu.cn.
\\
  \textsuperscript{*}Equal contribution
}

\begin{document}
\maketitle

\begin{abstract}

Financial multimodal reasoning requires agents to coordinate numerical computation, retrieval, visual interpretation, and temporal grounding across heterogeneous evidence sources. Existing tool-augmented agents improve execution fidelity, yet remain largely stateless across episodes, repeatedly rediscovering reasoning strategies and failure patterns. In high-stakes financial settings, this leads to unreliable tool routing, noisy retrieval, and hallucination-prone reasoning.
We present FinAcumen, a financial reasoning agent framework centered on selective experience memory for tool-augmented multimodal reasoning. FinAcumen accumulates financially grounded reasoning experience from prior trajectories, distilling successful strategies and failure-derived cautionary rules into a persistent memory bank. During inference, retrieved experiences condition reasoning only when semantic relevance exceeds a calibrated threshold, while irrelevant memory is explicitly suppressed through a fallback mechanism. A deterministic financial tool environment further grounds numerical computation, retrieval, visual decoding, and answer verification. Across four financial multimodal reasoning benchmarks, FinAcumen consistently improves a frozen 8B vision-language model over finance-specialized models and approaches leading proprietary general-purpose models. Further analysis shows that selective experience activation improves reasoning reliability under retrieval uncertainty. Our code is available at \url{https://github.com/CamelliaLilium/FinAcumen}

\end{abstract}

\section{Introduction}
\label{sec:intro}

Financial multimodal reasoning requires agents to coordinate heterogeneous evidence from text, tables, charts, and time-indexed records.
Unlike conventional question answering (QA), this setting jointly demands visual interpretation, numerical computation, retrieval, and temporal grounding across multiple interacting modalities.
Appendix~\ref{app:benchmark_examples} presents representative items from the four benchmarks evaluated in this work.
Recent benchmarks expose complementary aspects of this challenge, including quantity localization in filings of the United States Securities and Exchange Commission (SEC), chart-centric numerical reasoning, temporal-aware multimodal retrieval, and hallucination-sensitive financial analysis~\citep{bizbench,tang2025finmmr,zhu2025towards,luo2025finmme}.

Despite rapid progress in multimodal large language models (LLMs), reliable financial reasoning remains unresolved.
General-purpose models lack financially grounded reasoning strategies under complex multimodal settings, while finance-specialized models often exhibit limited robustness outside their tuning distributions~\citep{liu2025fin,huang2024open,caillaut2025llm}.
Recent evidence shows that even models performing competitively on distribution-near tasks can fail sharply on chart-dense and retrieval-intensive settings, suggesting that static fine-tuning alone does not provide adaptive reasoning policies for heterogeneous modality compositions~\citep{dai2026realfin,deng2026cler}.
Performance remains particularly constrained in high-difficulty settings, where benchmark leaders still exhibit substantial degradation under retrieval ambiguity, multi-step numerical reasoning, and hallucination-sensitive evaluation~\citep{tang2026finmmdocr,tang2025finmmr,luo2025finmme,zhu2025towards}.

In practical financial analysis workflows, human experts rely not only on domain knowledge, but also on accumulated experience about when to retrieve evidence, invoke tools, cross-check modalities, and verify intermediate conclusions.
Tool-augmented paradigms such as ReAct externalize computation and perception, yet remain largely stateless across episodes and repeatedly rediscover reasoning procedures~\citep{yao2023react}.
Existing memory-based paradigms partially address this limitation by storing prior reflections or trajectories, but they often conflate successful and failed experiences, retrieve weakly related guidance, and expand without controlling retrieval quality~\citep{shinn2023reflexion,zhao2024expel,ouyang2025reasoningbank,echosafe}.
Such behavior is particularly problematic in finance, where irrelevant retrieval can directly degrade reasoning quality and hallucinated evidence may lead to incorrect analytical conclusions~\citep{shi2023large,liu2024lost}.
Benchmarks such as FinTMMBench and FinMME make this issue explicit by penalizing retrieval errors and unsupported reasoning, requiring systems to determine not only what to retrieve, but also whether retrieved guidance should influence inference~\citep{zhu2025towards,luo2025finmme}.

We propose \textbf{FinAcumen}, a financial multimodal reasoning framework centered on selective experience-guided inference for tool-augmented agents.
FinAcumen introduces a \textbf{Financial Memory (FM)} module that accumulates financially grounded reasoning experience from prior trajectories, separating reusable strategies from failure-derived cautionary rules.
During inference, retrieved experience conditions reasoning only when semantic relevance exceeds a calibrated threshold; otherwise, the model explicitly falls back to its base reasoning policy to suppress noisy memory injection.
A complementary \textbf{Financial Tools (FT)} environment provides deterministic support for numerical computation, retrieval, visual decoding, and answer verification.

Our contributions are summarized as follows:
\begin{enumerate}
\item We introduce a selective experience memory framework for financial multimodal agents that separates reusable reasoning strategies from failure-derived cautionary guidance.
\item We propose a retrieval-conditioned inference mechanism in which memory activates only under sufficient semantic relevance, enabling fallback under retrieval uncertainty.
\item Across four financial multimodal reasoning benchmarks, FinAcumen consistently improves its frozen 8B vision-language model (VLM), exceeds the reported finance-specialized baselines on shared settings, and remains competitive with leading proprietary models.
\end{enumerate}

\section{Related Work}
\label{sec:related}

\paragraph{Financial multimodal reasoning.}
Financial QA evaluation has expanded from text-only settings to multimodal, temporal-aware, and hallucination-sensitive benchmarks. Representative datasets cover SEC-grounded quantity localization (BizBench, where we evaluate the SEC-NUM subset), chart-centric numerical reasoning (FinMMR), temporal multimodal retrieval and reasoning (FinTMMBench), and broad financial multimodal assessment (FinMME)~\citep{bizbench,tang2025finmmr,zhu2025towards,luo2025finmme}.
On the modeling side, finance-oriented LLMs such as Fin-R1~\citep{liu2025fin}, Open-FinLLMs~\citep{huang2024open}, and LLM Pro Finance Suite~\citep{caillaut2025llm} mainly inject domain knowledge through parameter updates.
In contrast, our formulation in \S\ref{sec:problem} and \S\ref{sec:method} keeps the backbone frozen and adapts at inference time through a fixed tool suite and retrieved experience, so performance differences are attributable to experience conditioning
rather than model re-training.

\paragraph{Cross-episode experience memory.}
Research on cross-episode agent improvement can be viewed along three steps: memory construction, retrieval decision, and bank maintenance. ReAct~\citep{yao2023react} established tool-interleaved reasoning but remains episode-local. Reflexion~\citep{shinn2023reflexion}, ExpeL~\citep{zhao2024expel}, ReasoningBank~\citep{ouyang2025reasoningbank}, and EchoSafe~\citep{echosafe} move to persistent experience accumulation across episodes. For retrieval, prior evidence shows that irrelevant context can harm reasoning quality~\citep{shi2023large,liu2024lost}, while Self-RAG~\citep{asai2024self} highlights explicit retrieve-or-not decisions.

Recent work also explores task-specific memory and skill artifacts. M$^\star$ evolves task-specific memory programs, whereas ASDA distills structured skills for financial reasoning without modifying model weights~\citep{pan2026m,yim2026asda}.

The use of a bank constructed before evaluation also follows established offline protocols. ExpeL distills insights from training tasks for use on unseen tasks~\citep{zhao2024expel}; Agent Workflow Memory induces reusable workflows offline from training trajectories~\citep{wang2025awm}; and MemGen constructs generative memory offline and keeps the memory pathway fixed during inference~\citep{zhang2026memgen}.
Relative to the cited approaches, FinAcumen combines three design choices: structured Findings and Cautions, an annotator that converts retrieved experience into directives for the current query, and calibrated thresholding with an empty-return fallback under a fixed tool suite.

\section{Problem Formulation}
\label{sec:problem}

\textbf{Notation.} We consider multimodal financial QA with disjoint train and test splits. Each instance $x = (q, c, I)$ comprises question $q$, textual context $c$, and optional images $I$, with gold answer $y^\star(x)$ scored under benchmark metric $\mu$. A frozen VLM $\pi_\theta$ performs multi-turn tool-augmented decoding over a fixed tool suite $\mathcal{T}$ (\S\ref{sec:method:ft}). A memory bank $\mathcal{M}$, populated from training trajectories, stores structured experience entries. At inference, entries are retrieved via embedding similarity and the top-$k$ are injected into the prompt (Algorithm~\ref{alg:solve}); a complete symbol table appears in Appendix~\ref{app:notation}.

Given $\mathcal{M}$ populated from $\mathcal{D}_{\text{train}}$ trajectories, we seek $\mathcal{M}^\star$ that maximizes expected accuracy on $\mathcal{D}_{\text{train}}$:
\begin{equation}
  \mathcal{M}^\star \in \arg\max_{\mathcal{M}} \;
  \mathbb{E}_{x \sim \mathcal{D}_{\text{train}}}
  \Bigl[\,
    \mu\bigl(\pi_\theta(x; \mathcal{M}, \mathcal{T}),\, y^{\star}(x)\bigr)
  \Bigr].
  \label{eq:obj}
\end{equation}
At test time, $\mathcal{M}^\star$ is frozen and read-only: queries may activate stored
patterns but never write to the bank. Freezing the bank removes a potential leakage path in which gold-answer feedback from one test item is written to memory and later retrieved by another, thereby preserving the intended independence and reproducibility of held-out evaluation. Final performance is reported as
$\mathbb{E}_{x\sim\mathcal{D}_{\text{test}}}[\mu(\pi_\theta(x;\mathcal{M}^\star,\mathcal{T}), y^\star(x))]$.

The discrete optimization in \eqref{eq:obj} is approximated in
Section~\ref{sec:method} through a writer-retriever loop (Algorithm~\ref{alg:solve})
that constructs $\mathcal{M}$ from graded training trajectories.

\section{Method}
\label{sec:method}

\subsection{Overview}
\label{sec:method:overview}

\begin{figure*}[t]
  \centering
  \includegraphics[width=\textwidth]{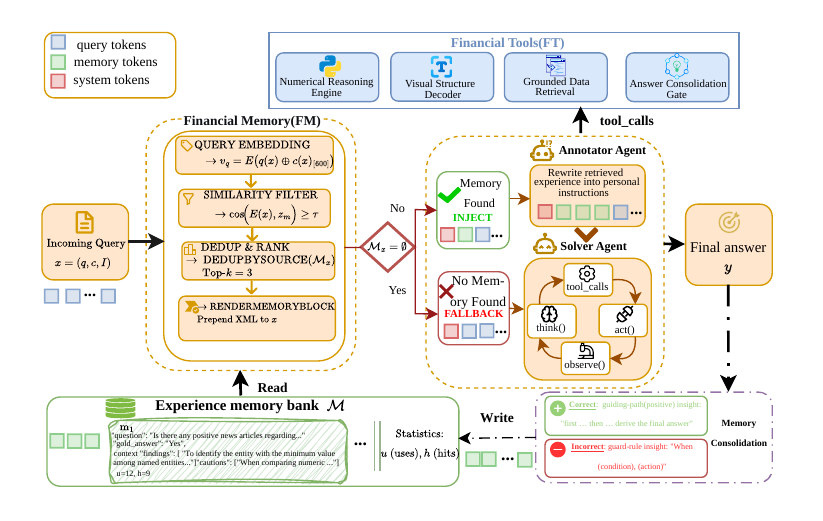}
  \caption{The \fa{} pipeline. Experience accumulated through
    multi-trajectory sampling is stored in \fm{} as generalized strategies
    and guard rules. At inference, semantically similar entries are
    retrieved and provided as in-context guidance; when none qualify, the
    model proceeds with \ft{} alone.}
  \label{fig:finacumen_overview}
\end{figure*}

As defined in Section~\ref{sec:problem}, a frozen language model
$\pi_\theta$ operates over a tool suite $\mathcal{T}$ to solve financial
problems.
\fa{} extends this formulation with two components, illustrated in
Figure~\ref{fig:finacumen_overview}: Financial Memory (\fm{}) and Financial
Tools (\ft{}).
\fm{} enables the model to accumulate experience from its own solution
trajectories and retrieve it as guidance for new problems, growing a bank
of generalized strategies and guard rules through multi-trajectory
sampling.
\ft{} provides a deterministic execution layer that offloads computation,
data retrieval, visual decoding, and answer consolidation from the
generative process.
The pipeline is shared across accumulation and inference, isolating
memory retrieval as the sole conditioned pathway.

\subsection{Financial Memory}
\label{sec:method:fm}

\paragraph{Memory bank structure.}
Each entry in $\mathcal{M}$ is a structured record of one completed
problem, containing the original question, its gold answer, and the
distilled experience.
The experience consists of two parts: strategies generalized from
trajectories that reached the correct answer, and guard rules extracted
from those that did not.

\paragraph{Memory consolidation.}
For each problem the model encounters, the framework generates multiple
independent solution trajectories from $\pi_\theta$ with diverse decoding,
scores each against the gold answer, and feeds the scored set to a single
summary agent.
Let $\{ \mathrm{Traj}_1, \dots, \mathrm{Traj}_K \}$ denote the trajectories for problem $x$
with gold answer $y^*$, and let $s_k = \text{Score}(\mathrm{Traj}_k, y^*)$ be the
correctness score assigned to each.
The summary agent synthesizes two outputs:
\begin{equation}
  (S, C) = \text{Summarize}\bigl( \{ (\mathrm{Traj}_k, s_k) \}_{k=1}^{K}, \; y^* \bigr),
  \label{eq:consolidate}
\end{equation}
where $S$ collects strategies from successful trajectories and $C$
collects cautions from failed ones.
These are written into a new entry alongside the original question and
answer, incrementally expanding the bank for future retrieval.

\paragraph{Memory activation.}
At inference, a query $x$ is embedded into the same semantic space as the
bank through the shared encoder $E(\cdot)$.
Its similarity to each entry $m$ is defined by cosine distance over the
resulting vectors:
\begin{equation}
  \text{sim}(x, m) = \frac{E(x) \cdot z_m}{\|E(x)\| \, \|z_m\|},
  \label{eq:sim}
\end{equation}
where $z_m$ denotes the pre-stored embedding of entry $m$.
Entries passing a calibrated threshold $\tau$ form the candidate set
\begin{equation}
  \mathcal{M}_x = \{\, m \in \mathcal{M} \mid \text{sim}(x, m) \ge \tau \,\},
  \label{eq:threshold}
\end{equation}
which is then deduplicated and ranked.
The top-ranked subset $\mathcal{M}_x^*$ is rendered as a structured prefix in Extensible Markup Language (XML) and prepended before the problem prompt.
The model produces its final answer $y$ through a tool-augmented rollout
\begin{equation}
  y =
  \begin{cases}
    \text{Rollout}\bigl( \pi_\theta,\; x_{\text{mem}},\; \mathcal{T} \bigr), & \mathcal{M}_x^* \ne \emptyset \\[4pt]
    \text{Rollout}\bigl( \pi_\theta,\; x,\; \mathcal{T} \bigr),              & \text{otherwise},
  \end{cases}
  \label{eq:rollout}
\end{equation}
where $x_{\text{mem}}$ denotes the prompt with memory prefix prepended.

\subsection{Agent Architecture}
\label{sec:method:agent}

All LLM-driven roles in \fa{} share the same base model and are
distinguished by their prompt-defined behavior.
During consolidation, the model acts as a sampling agent that
generates multiple diverse solution trajectories, and as a summary
agent that scores them against the gold answer before synthesizing
strategies from successes and cautions from failures.
During inference, an annotator agent rewrites retrieved experiences
into directives personalized to the current problem before the
solver agent plans and executes tool-augmented reasoning.
The solver internally decomposes into two phases: planning, where it
inventories inputs, judges experience applicability, and formulates
a strategy; and execution, where it invokes tools and terminates
with the final answer.

\subsection{Financial Tools}
\label{sec:method:ft}

The tool suite $\mathcal{T}$ comprises four tools:
\begin{equation}
\mathcal{T} = \left\{
  \begin{gathered}
    \text{Numerical Reasoning Engine},\\
    \text{Grounded Data Retrieval},\\
    \text{Visual Structure Decoder},\\
    \text{Answer Consolidation Gate}
  \end{gathered}
\right\},
\label{eq:tools}
\end{equation}
covering precise arithmetic, factual data lookup, chart interpretation,
and answer verification.
The same tool set and calling schema are shared across all configurations,
so that tool mechanics remain identical whether or not \fm{} conditions the
prompt.

\paragraph{Numerical Reasoning Engine.}
This tool decouples financial arithmetic from the generative process by
executing code in a persistent namespace.
Let $\sigma_t$ denote the variable bindings accumulated across consecutive
calls within the same agent loop, with $\sigma_0$ containing no bindings:
\begin{equation}
  (o_t, \sigma_t) = \textsc{Execute}(\text{code}_t,\; \sigma_{t-1}),
  \label{eq:engine}
\end{equation}
where $o_t$ is the captured output.
This stateful design enables cascaded operations such as rate compounding
and cross-statement aggregation that stateless code generators cannot
reliably chain: without persistent state, the model must reinitialize all
quantities in every code block, risking inconsistency across arithmetic
steps.
Financial multi-step reasoning chains intermediate quantities by nature:
growth rates build on prior period values, ratios depend on previously
computed numerators and denominators.
A fresh execution session per code call severs this dependency chain; the
persistent namespace preserves it.

\paragraph{Grounded Data Retrieval.}
This tool resolves factual financial queries against a curated panel of
open, high, low, and close (OHLC) price records, news corpora, and quarterly indicator tables.
Given a symbol, date, and indicator, the lookup returns the corresponding
structured entry:
\begin{equation}
  r = \textsc{Lookup}(\text{symbol},\; \text{date},\; \text{indicator}),
  \label{eq:lookup}
\end{equation}
anchoring each numerical claim to a verifiable source row.
All filtering logic executes deterministically within the tool backend,
preventing the model from fabricating lookup parameters or results.
Financial quantities such as quarterly revenue or daily closing prices are
factual, time-bound values.
Parametric knowledge, encoded during pre-training, cannot reliably reproduce
them when the query demands a specific date or indicator; consulting a
deterministic panel anchors each claim in an externally verifiable source.

\paragraph{Visual Structure Decoder.}
This tool handles chart and table images in two layers.
Before the agent loop begins, all available images are transcribed through optical character recognition (OCR)
and prepended to the context as structured text, providing precise
numerical inputs upfront.
During reasoning, if the agent detects that computation results contradict
values previously read from images, it may explicitly invoke this tool
to re-parse the source image, producing an updated structured
representation that resolves the discrepancy.

\paragraph{Answer Consolidation Gate.}
This tool serves as the sole termination point of the agent loop.
Given the accumulated trajectory $\mathrm{Traj}$ and the proposed answer $a$, it
enforces three conditions before emission:
\begin{equation}
  \begin{aligned}
    \textsc{Verify}(\mathrm{Traj}, a) \; \equiv \;
    &\text{source-traceable}(\mathrm{Traj}) \\
    \wedge\;&\text{units-ok}(\mathrm{Traj}) \\
    \wedge\;&\text{format-valid}(a),
  \end{aligned}
  \label{eq:verify}
\end{equation}
where source-traceable checks that $a$ or its numeric equivalent appears
among the captured outputs of Execute calls, the returned entries of
Lookup calls, or the OCR-extracted values in the context; units-ok verifies
that quantity units remain compatible across tool calls without unresolved
conversions; and format-valid ensures $a$ conforms to the task instruction.
Because termination is already the final reasoning step, embedding the
verification check there costs no additional inference; a separate
post-hoc auditor would require a second traversal of the full trajectory.

\paragraph{Interface with \fm{}.}
\fm{} is the memory bank described in \S\ref{sec:method:fm};
$\pi_\theta$ is the frozen language model.
\fm{} operates strictly at the prompt level and never interleaves with
the tool-calling mechanics.
The baseline executes $\textsc{Rollout}(\pi_\theta, x, \mathcal{T})$ without
memory, while the memory-active path prepends retrieved experiences before
$x$.
Because $\mathcal{T}$ and the calling protocol remain identical across both
paths, any performance difference is attributable solely to the experiential
guidance rather than altered tool dynamics.

\section{Experiments}
\label{sec:exp}

\subsection{Experimental Setup}
\label{sec:exp:setup}

\paragraph{Benchmarks.}
We evaluate six settings from four benchmarks, covering numerical grounding, multimodal numerical reasoning, temporal-aware retrieval, and chart understanding.
BizBench SEC-NUM \citep{bizbench} includes 2000 SEC 10-K/10-Q items for quantity localization under numeric distractors; we use exact match with 1\% relative tolerance.
FinMMR \citep{tang2025finmmr} includes Easy/Medium/Hard splits (1200/1200/1000) for bilingual multi-image financial numerical reasoning; we follow its 0.2\% relative tolerance criterion.
FinTMMBench \citep{zhu2025towards} includes 1136 temporal-aware multimodal retrieval-reasoning items over tables, news, prices, and charts; we report official GPT-4o-mini-judge accuracy.
FinMME \citep{luo2025finmme} includes 2220 chart-intensive items across diverse financial domains. We report Avg as the arithmetic mean of the six dimensions defined by the benchmark. Because the released annotations do not provide official per-sample domain labels, we reconstruct the corresponding dimension partition as documented in Appendix~\ref{app:finmme_metric}.
We follow each benchmark's official split and evaluation protocol.
Full dataset, scoring, and implementation details are provided in Appendix~\ref{app:impl}.
Representative benchmark items are provided in Appendix~\ref{app:benchmark_examples}.

\paragraph{Implementation details.}
We use frozen Qwen3-VL-8B-Instruct~\citep{qwen3vl} through the DashScope application programming interface (API) with temperature 0. Without augmentation, it serves as the \textsc{Base} model-only baseline in Table~\ref{tab:main}.
As detailed in Section~\ref{sec:method:ft}, the tool suite (\textsc{Base} + \ft{}) equips the backbone with a ReAct loop (max~16~steps) over four tool components for numerical reasoning, data retrieval, visual decoding, and answer gating; its standalone contribution is isolated in the ablation study (Section~\ref{sec:exp:ablation}).
The \fm{} bank pools $\sim$2,400 experience accumulation trajectories across all benchmarks, sourced exclusively from training\slash validation splits held out from test data, and is frozen read-only at evaluation.
All ablation conditions share identical tool and decoding configurations, isolating the effect of \fm{} and \ft{}.
Published baseline numbers are collected from respective benchmark sources; exact model variant specifications and per-cell attributions appear in Appendix~\ref{app:baselines}, with further implementation details in Appendix~\ref{app:impl}. Construction and inference cost estimates appear in Appendix~\ref{app:cost}.

\subsection{Main Results}
\label{sec:exp:main}

Table~\ref{tab:main} presents a system-level comparison under a constrained deployment setting. In this work, constrained deployment refers to inference-time augmentation of a selected frozen backbone with external memory and deterministic tools while preserving the backbone parameters and deployment interface. The selected backbone may be accessed through an API or deployed locally. Published closed-source baselines serve as benchmark-reported capability references and do not use \fm{} or \ft{}. Accordingly, the controlled comparison is between Base and \fa{} under the same backbone and evaluation protocol.

The difference between Base and \fa{} estimates the framework contribution under the shared backbone, while the best reported General LLM result indicates the remaining capability gap under each benchmark setting. Model variant specifics are documented in Appendix~\ref{app:baselines}.

\begin{table*}[t]
  \caption{Per-benchmark accuracy across six financial QA settings. Published baseline entries are benchmark-reported model results without our \fm{} and \ft{} wrapper; per-cell variant details and source attributions are in Appendix~\ref{app:baselines}. Best result per column in bold, second-best underlined. A dash indicates unreported by the source benchmark. \fa{} results use a frozen Qwen3-VL-8B-Instruct backbone with \fm{} and \ft{}.}
   \label{tab:main}
  \centering
  \small
  \setlength{\tabcolsep}{5pt}
  \begin{tabular}{@{}llcccccc@{}}
    \toprule
    \multirow{2}{*}{Group} & \multirow{2}{*}{Model} & \multicolumn{1}{c}{BizBench} & \multicolumn{3}{c}{FinMMR} & \multirow{2}{*}{FinTMMBench} & \multirow{2}{*}{FinMME} \\
    \cmidrule(lr){3-3}\cmidrule(lr){4-6}
    & & SEC-NUM & Easy & Medium & Hard & & \\
    \midrule
    \multirow{5}{*}{General LLMs} 
    & GPT         & \textbf{79.30} & \underline{78.00} & 63.33 & 45.40  & \textbf{21.53}  & 46.56 \\
    & Claude Sonnet & 63.39         & 77.00          & 62.25          & \textbf{50.80} & --     & 48.20 \\
    & Qwen2.5-VL-72B & 69.85       & 77.42          & \underline{63.42}          & 43.30  & 15.15  & \textbf{52.54} \\
    & Llama         & \underline{74.70}         & 77.83          & 63.25          & \underline{48.70}  & 11.57  & -- \\
    & Gemini~2.0 Flash & --         & 74.92          & 57.83          & 44.40  & 18.42  & \underline{51.85} \\
    \midrule
    \multirow{3}{*}{Financial LLMs}
    & Llama-Open-Finance-8B & 63.64 & -- & -- & -- & --     & 29.95 \\
    & FinLLaVA              & 53.46 & 3.25 & 2.75 & 3.90 & 11.80 & 27.70 \\
    & Fin-R1                & 59.32 & --   & --   & --   & --    & 26.62 \\
    \midrule
    \multirow{1}{*}{Base}
    & Qwen3-VL-8B-Instruct   & 27.67 & 59.17 & 45.75 & 31.40 & 16.20 & 27.84 \\
    \midrule
    \multirow{1}{*}{\fa{}}
    & Qwen3-VL-8B + \fa{}    & 68.65 & \textbf{81.67} & \textbf{67.00} & 46.10 & \underline{19.98} & 51.08 \\
    \bottomrule
  \end{tabular}
\end{table*}

Table~\ref{tab:main} supports three observations.
\textbf{First, \fa{} produces large system-level gains}: the same frozen backbone rises from 27.67 to 68.65 on BizBench and from 27.84 to 51.08 on the FinMME Avg metric, while also surpassing the published GPT-4o entries on FinMMR Easy/Medium.
\textbf{Second, it is more robust than static financial specialization}: across all shared columns, \fa{} outperforms Llama-Open-Finance-8B, FinLLaVA, and Fin-R1, suggesting that financial multimodal QA benefits from test-time evidence routing, verification, and strategy adaptation.
\textbf{Third, the remaining gaps are informative}: \fa{} still trails the best General LLM on BizBench, FinMMR Hard, and FinTMMBench, where stronger planning, temporal retrieval, or difficult numerical reasoning is required.
Thus, \fa{} is best viewed as a deployment-efficient way to strengthen frozen VLMs, not as evidence that an 8B backbone has superior intrinsic parametric ability.
Repeated-run uncertainty, experience-memory baseline comparisons, and cross-backbone evaluations are reported in Appendices~\ref{app:uncertainty}, \ref{app:memory_baselines}, and~\ref{app:cross_backbone}.

\subsection{Ablation Study}
\label{sec:exp:ablation}

Since Table~\ref{tab:main} compares a full system against published model-only baselines, we isolate the source of the gain by toggling \ft{} and \fm{} on the same Qwen3-VL-8B backbone.
Table~\ref{tab:ablation} reports FinMMR Easy, whose moderate base accuracy (59.17) makes module-level gains easy to measure.
All variants share the same backbone, decoding configuration, and scoring protocol.

\begin{table}[t]
  \caption{Ablation of \ft{} and \fm{} on FinMMR Easy with Qwen3-VL-8B-Instruct. $\Delta$ denotes absolute improvement over the Base only variant.}
  \label{tab:ablation}
  \centering
  \small
  \begin{tabular}{@{}lcc@{}}
    \toprule
    Variant & Accuracy & $\Delta$ \\
    \midrule
    Base only                 & 59.17 & -- \\
    Base + \ft{}              & 74.08 & +14.91 \\
    Base + \ft{} + \fm{} (full \fa{}) & 81.67 & +22.50 \\
    \bottomrule
  \end{tabular}
\end{table}

The ablation explains the main-table gains.
Adding \ft{} raises accuracy from 59.17 to 74.08, showing that tools resolve many execution-level errors.
Adding \fm{} on top of the same tool suite further improves accuracy to 81.67, a 7.59-point gain over tool-only inference.
This shows that \fa{} is not merely a tool wrapper: retrieved experiences provide strategy-level guidance about tool use, quantity verification, and recurring financial reasoning traps.
Together, \ft{} and \fm{} yield a 22.50-point improvement over the base model.
The larger-backbone component study in Appendix~\ref{app:cross_backbone} shows a similar complementary pattern.

\subsection{Memory Accumulation Behavior}
\label{sec:exp:curve}

To isolate the marginal contribution of experience memory, we evaluate a frozen \fm{} bank on BizBench SEC-NUM across seven snapshots spanning 0 to 2400 accumulated entries. The $n=0$ condition corresponds to the \ft{}-augmented agent with an empty \fm{} bank and is distinct from the bare model baseline.

\begin{figure}[t]
  \centering
  \includegraphics[width=\linewidth]{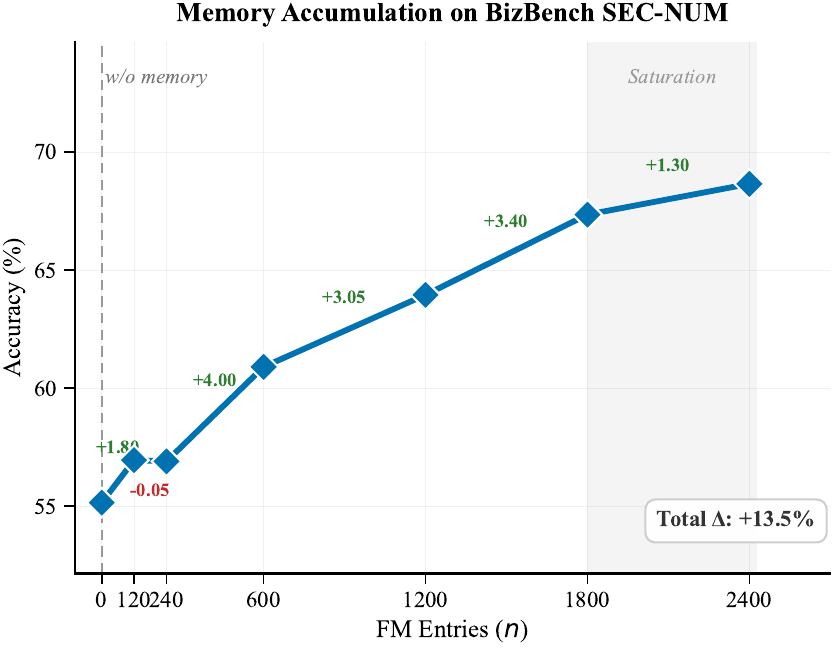}
  \caption{Accuracy on BizBench SEC-NUM as a function of accumulated \fm{} bank size. The $n=0$ condition includes the \ft{} tool suite. Per-step delta annotations reveal diminishing marginal returns, with accuracy approaching a saturation regime beyond $n=1800$, marked by the shaded region.}
  \label{fig:curve}
\end{figure}

Figure~\ref{fig:curve} reports the accuracy trajectory. Starting from 55.15\% with no experience entries, accuracy rises to 68.65\% at 2400 entries, a net gain of 13.5 points. The curve exhibits diminishing returns: per-interval gains are most pronounced at smaller bank sizes and progressively decline, entering a saturation regime beyond 1800 entries where additional entries produce marginal improvements. A detailed decomposition of per-interval marginal gains is provided in Appendix~\ref{app:saturation}. Across the evaluated snapshots, accuracy increases overall as the bank expands to 2400 entries, while the smaller gains at larger bank sizes indicate diminishing marginal returns. This analysis characterizes the relationship between bank size and test accuracy under the fixed retrieval setting; it does not by itself isolate the effect of selective retrieval.

\subsection{Sensitivity Analysis}
\label{sec:exp:sensitivity}

The \fm{} retrieval pipeline has two hyperparameters: a similarity threshold $\tau$ governing entry admission and a retrieval depth $k_{\text{max}}$ governing how many admitted entries enter the reasoning context. We analyze each independently.

\begin{figure*}[t]
  \centering
  \begin{subfigure}[b]{0.48\textwidth}
    \centering
    \includegraphics[width=\linewidth]{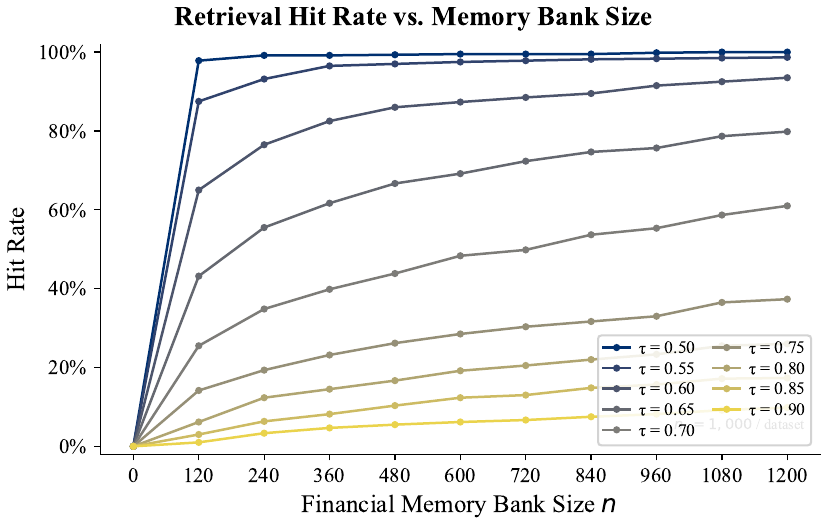}
    \caption{Hit rate as a function of bank size, stratified by similarity threshold $\tau$.}
    \label{fig:hitrate}
  \end{subfigure}
  \hfill
  \begin{subfigure}[b]{0.48\textwidth}
    \centering
    \includegraphics[width=\linewidth]{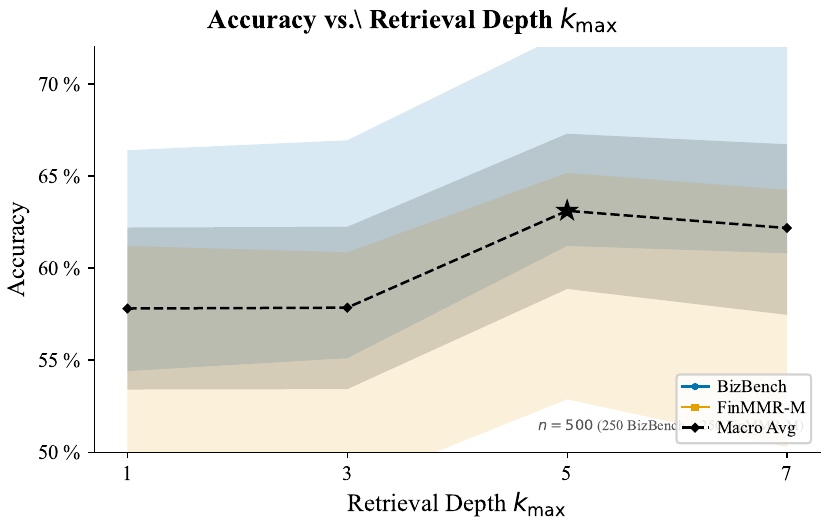}
    \caption{Accuracy as a function of retrieval depth $k_{\text{max}}$.}
    \label{fig:kmax_sensitivity}
  \end{subfigure}
  \caption{Sensitivity analysis of two retrieval hyperparameters. (a) Hit rate across bank sizes for different cosine thresholds; three operational regimes emerge, with the practical range centered at moderate filtering strictness. (b) Accuracy as a function of retrieval depth, following an inverted-U pattern that peaks at moderate depth and degrades at both extremes. Shaded regions denote 95\% bootstrap confidence intervals.}
  \label{fig:sensitivity}
\end{figure*}

\paragraph{Cosine similarity threshold $\tau$.}
We evaluate $\tau$ via hit rate rather than downstream accuracy, as this threshold must establish a viable candidate pool before $k_{\text{max}}$ can operate; hit rate measures the fraction of test queries for which at least one bank entry exceeds the threshold. Figure~\ref{fig:hitrate} shows hit rates for nine $\tau$ values spanning $[0.50, 0.90]$ against accumulated bank size. Three operational regimes emerge. A loose-filtering regime at $\tau \leq 0.55$ exceeds 87\% hit rate at 120 entries with negligible semantic filtering. A strict-blocking regime at $\tau \geq 0.75$ stays below 37\% even at 1200 entries, denying most queries access to memory. A moderate-filtering regime in $[0.60, 0.70]$ yields 43\% to 80\% across bank expansion. At $\tau = 0.65$, the center of this moderate regime, hit rate rises from 43.2\% at 120 entries to 79.8\% at 1200 entries; saturation beyond 600 entries mirrors the accumulation plateau observed in Figure~\ref{fig:curve}, consistent with early entries capturing prevalent error patterns. We therefore select $\tau=0.65$ as a balance between retrieval coverage and filtering strength and keep this value fixed across benchmarks to avoid tuning for individual benchmarks.

\paragraph{Retrieval depth $k_{\text{max}}$.}
We test $k_{\text{max}} \in \{1, 3, 5, 7\}$ at $\tau = 0.65$ on 500 questions evenly split across BizBench and FinMMR-M, with all other configurations matching the main experiments. Figure~\ref{fig:kmax_sensitivity} shows an inverted-U relationship between retrieval depth and accuracy. At $k_{\text{max}} = 1$ macro accuracy is 57.80\%; expanding to $3$ reaches 57.85\%, a shift of 0.05 percentage points that lies within bootstrap confidence intervals and is thus not statistically meaningful. At $k_{\text{max}} = 5$ macro accuracy rises to 63.11\%, a gain of 5.26 points that falls well outside the confidence bounds, with consistent improvements on BizBench of 5.98 points and FinMMR-M of 4.55 points. This sharp transition is consistent with a scenario where entries ranked fourth and fifth cover strategy categories absent from the top three; the highest-ranked entries tend to cluster around a single dominant error type, while moderately-ranked entries span complementary failure modes demanding qualitatively different solution patterns. At $k_{\text{max}} = 7$ macro accuracy drops to 62.17\%, a decline of 0.94 points, indicating that retrieval beyond five entries introduces distractors that outweigh any additional strategic signal.
The tension between insufficient strategic coverage at low $k_{\text{max}}$ and signal dilution at high $k_{\text{max}}$ is resolved at $k_{\text{max}} = 5$, which is adopted for all experiments.

\section{Conclusion}
\label{sec:conclusion}

We introduced \fa{}, a framework coupling self-evolving Financial Memory
with deterministic Financial Tools for financial multimodal QA.
Across four benchmarks, \fa{} improves its frozen 8B backbone, exceeds
the reported finance-specialized baselines on shared settings, and remains
competitive with leading general-purpose systems.
Ablation confirms complementary roles: \fm{} improves routing and
strategy selection via polarity-aware memory activation, while \ft{}
guarantees numerical fidelity through execution-environment grounding;
when no relevant experience is retrievable, \fm{} falls back to the
tool-augmented baseline without degradation.
The decoupled architecture, in which an evolving experience bank operates
orthogonally to a deterministic tool suite around a frozen base model,
suggests a general paradigm for artificial intelligence systems in professional domains: externalized
memory captures reusable strategy patterns without destabilizing the
core model, while domain-specific tools provide computational guarantees
that generative components alone typically struggle to guarantee.
Future work includes sample-efficient consolidation and compositional
memory to reduce cold-start cost and support cross-task transfer,
together with cross-lingual alignment and stronger visual encoders for
multilingual and chart-dense deployment settings.
\section{Limitations}
\label{sec:limitations}

We acknowledge two primary limitations. First, Financial Memory construction requires auxiliary LLM calls for trajectory tagging, embedding, and experience consolidation, introducing a non-trivial upfront cost during bank initialization. Although retrieval remains efficient after consolidation, improving sample efficiency and enabling incremental online accumulation remain important future directions. Second, our evaluation is limited to English financial multimodal benchmarks. Financial documents vary substantially across languages in terminology, tabular conventions, and chart structures, which may limit direct transferability of retrieved experience and tool configurations.
Extending experience-conditioned reasoning to multilingual financial settings and broader model families remains future work.

\section*{Acknowledgments}
This work was supported in part by the National Natural Science Foundation of China under Grant 62172053 and Grant 62302059.

\bibliography{references}

\appendix
\newpage
\noindent{\Large\textbf{Appendix}}

\vspace{1.5em}

\noindent\textbf{A\quad Notation}
\dotfill \pageref{app:notation}

\noindent\textbf{B\quad Inference Algorithm}
\dotfill \pageref{app:algorithm}

\noindent\textbf{C\quad Marginal Gain Decomposition of Memory Accumulation}
\dotfill \pageref{app:saturation}

\noindent\textbf{D\quad Memory Strategy Ablation}
\dotfill \pageref{app:strategy}

\noindent\textbf{E\quad Retrieval Stability Analysis}
\dotfill \pageref{app:retrieval_stability}

\noindent\qquad\textit{E.1 Experimental Design}
\dotfill \pageref{sec:retrieval_exp_design}

\noindent\qquad\textit{E.2 Retrieval Metrics}
\dotfill \pageref{sec:retrieval_metrics}

\noindent\qquad\textit{E.3 Coverage and Retrieval Density}
\dotfill \pageref{sec:retrieval_coverage}

\noindent\qquad\textit{E.4 Retrieval Stability}
\dotfill \pageref{sec:retrieval_stability_analysis}

\noindent\textbf{F\quad Benchmark Examples}
\dotfill \pageref{app:benchmark_examples}

\noindent\textbf{G\quad Experimental Details}
\dotfill \pageref{app:impl}

\noindent\qquad\textit{G.1 Evaluation Protocols}
\dotfill \pageref{app:eval_protocols}

\noindent\qquad\textit{G.2 FinAcumen Configuration}
\dotfill \pageref{app:finacumen_config}

\noindent\qquad\textit{G.3 Baseline Attribution}
\dotfill \pageref{app:baselines}

\noindent\textbf{H\quad Repeated-Run Uncertainty}
\dotfill \pageref{app:uncertainty}

\noindent\textbf{I\quad Experience-Memory Baselines}
\dotfill \pageref{app:memory_baselines}

\noindent\textbf{J\quad Cross-Backbone and Scale Generalization}
\dotfill \pageref{app:cross_backbone}

\noindent\textbf{K\quad Cost Analysis}
\dotfill \pageref{app:cost}

\noindent\textbf{L\quad Prompt Templates}
\dotfill \pageref{app:prompts}

\section{Notation}
\label{app:notation}

We define mathematical symbols that appear across multiple sections of the main text. Symbols introduced and used only within a single paragraph or algorithm are defined locally and not repeated here.

\begin{table}[htbp]
  \centering
  \small
  \caption{Global mathematical notation.}
  \label{tab:notation}
  \begin{tabular}{@{}>{\raggedright\arraybackslash}p{0.30\linewidth}>{\raggedright\arraybackslash}p{0.65\linewidth}@{}}
    \toprule
    \textbf{Symbol} & \textbf{Definition} \\
    \midrule
    \multicolumn{2}{@{}l}{\textit{Problem and model}} \\
    \midrule
    $x=(q,c,I)$ & Instance: question $q$, text context $c$, optional images $I$ \\
    $\pi_\theta$ & Frozen vision-language model with parameters $\theta$ \\
    $\mathcal{T}$ & Financial tool suite (\S\ref{sec:method:ft}) \\
    \midrule
    \multicolumn{2}{@{}l}{\textit{Memory bank and retrieval}} \\
    \midrule
    $\mathcal{M}$ & Experience memory bank \\
    $z_m$ & Pre-computed embedding of entry $m\in\mathcal{M}$ \\
    $E(\cdot)$ & Unit-scale text embedding operator \\
    $\tau$ & Cosine similarity threshold ($\tau=0.65$) \\
    \bottomrule
  \end{tabular}
\end{table}

\section{Inference Algorithm}
\label{app:algorithm}
\begin{algorithm}[t]
  \caption{\fa{} single-instance inference with retrieval}
  \label{alg:solve}
  \begin{algorithmic}[1]
    \raggedright
    \Require $x$, $\mathcal{M}$, $\pi_\theta$, $\mathcal{T}$, $\omega$, $E_{\mathrm{mgr}}$
    \State $v_q \gets E(q(x) \oplus c(x)_{[:600]})$
    \State $V_{\mathcal{M}} \gets E_{\mathrm{mgr}}.\text{resolve\_all}(\mathcal{M})$
    \State $\vec{s} \gets v_q \cdot V_{\mathcal{M}}^{\top}$
    \State $I \gets \{i \mid s_i \ge 0.65\}$
    \If{$I = \emptyset$}
      \State \Return $\textsc{Rollout}(\pi_\theta, x, \mathcal{T})$
    \EndIf
    \State $\vec{I} \gets \textsc{DescendingSort}(I, \vec{s})$
    \State $\vec{I} \gets \textsc{DedupBySource}(\vec{I})$
    \State $\vec{I}_{\text{top}} \gets \vec{I}_{[:5]}$
    \State $\text{xml} \gets \textsc{RenderMemoryBlock}(\mathcal{M}[\vec{I}_{\text{top}}])$
    \State $y \gets \textsc{Rollout}\!\big(\pi_\theta, \textsc{Prepend}(\text{xml}, x), \mathcal{T}\big)$
    \State $\textsc{UpdateStats}(\vec{I}_{\text{top}}, y)$
    \If{$\omega$}
      \State \textbf{schedule async} $\textsc{Consolidate}(x, y, \mathcal{M})$
    \EndIf
    \State \Return $y$
  \end{algorithmic}
\end{algorithm}

\section{Marginal Gain Decomposition of Memory Accumulation}
\label{app:saturation}

\begin{figure}[t]
  \centering
  \includegraphics[width=\linewidth]{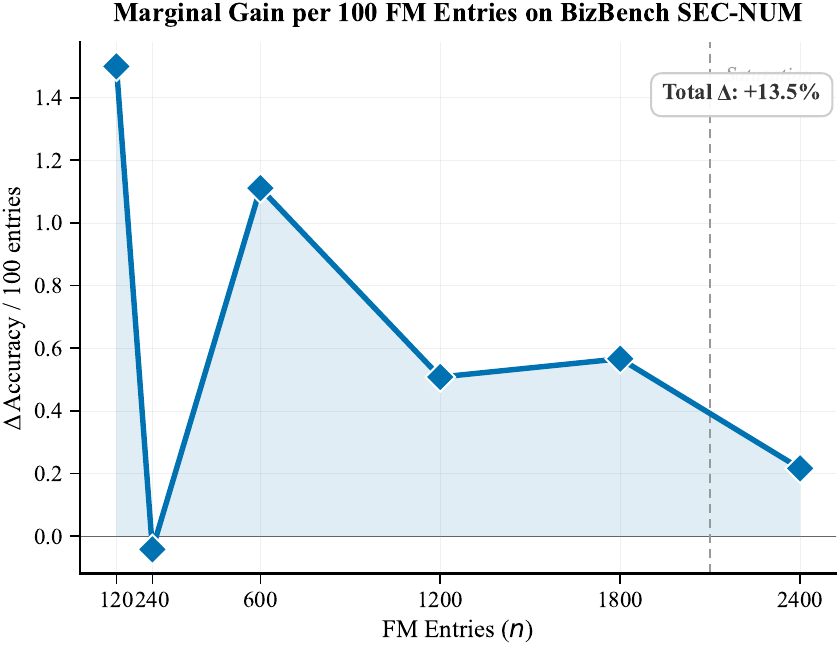}
  \caption{Marginal accuracy gain per 100 accumulated \fm{} entries on BizBench SEC-NUM. The curve traces a diminishing-returns pattern across consecutive snapshot intervals. The dashed vertical line separates the main accumulation phase from the saturation regime.}
  \label{fig:saturation_appendix}
\end{figure}

Figure~\ref{fig:saturation_appendix} decomposes the memory accumulation curve from Section~\ref{sec:exp:curve} into per-segment marginal gains, defined as the accuracy improvement per 100 accumulated \fm{} entries for each consecutive snapshot interval. The evaluation protocol mirrors the main analysis: the \ft{}-augmented Qwen3-VL-8B-Instruct backbone is evaluated on the full 2000-item BizBench SEC-NUM test set at seven bank snapshots spanning 0 to 2400 entries.

The marginal gain decomposition in Figure~\ref{fig:saturation_appendix} reveals a three-phase accumulation dynamic on BizBench.
\textbf{Phase 1 (0--120 entries):} The initial accumulation of 120 entries yields a marginal gain of +1.50 per 100 entries, capturing the most common error patterns.
\textbf{Phase 2 (240--1800 entries):} The next 1560 entries sustain a robust marginal efficiency of +0.51 to +1.11 per 100 entries, with the 240--600 interval producing the strongest per-entry return (+1.11) and the 600--1200 interval yielding +0.51. The 1200--1800 interval yields +0.57 per 100 entries, reflecting an uptick in marginal efficiency as newly accumulated entries target previously uncovered error patterns.
\textbf{Phase 3 ($n \ge 1800$):} Beyond 1800 entries, the marginal gain declines to +0.22 per 100 entries, or approximately +1.30 points across the final 600-entry block. This diminishing return supports the selection of 2400 entries as the saturation knee for the production bank.

The consistently positive marginal gains across all intervals, including the saturation phase, confirm that the \fm{} bank continues to provide useful signal even after substantial accumulation. While individual snapshots contain minor fluctuations (the 120--240 interval shows a near-zero change of $-0.04$), the overall trajectory is monotonically increasing on the macro timescale. This characteristic supports our deployment strategy of accumulating experience passively throughout training and freezing the bank at the empirically determined saturation point for test-time inference, as described in Section~\ref{sec:exp:curve}.

\section{Memory Strategy Ablation}
\label{app:strategy}

This appendix isolates the marginal contribution of each component in the \fa{} pipeline by comparing five injection strategies that toggle four fields of the retrieved memory block.

\begin{itemize}[nosep,leftmargin=*]
  \item \textbf{Q} (Question): the original question text of the top-retrieved similar item.
  \item \textbf{A} (Answer): the gold answer of the retrieved item.
  \item \textbf{E} (Experience): distilled Findings, reusable solution recipes, and Cautions, guard rules from past failures.
  \item \textbf{Annot} (Annotation): LLM-generated personalized directives that map generic rules to query-specific entities such as ticker symbols, date ranges, and metric names.
\end{itemize}

All strategies share Qwen3-VL-8B-Instruct as the backbone and operate on the same 600-question test subsample with MIN\_COSINE set to 0.60. Strategies A and B use the single-shot baseline-raw variant without tools; strategies C through E use the full FinAcumen variant with the four-tool ReAct loop. Table~\ref{tab:strategy} reports per-benchmark accuracy and the macro average across the four benchmarks.

\begin{table*}[t]
  \caption{Memory strategy ablation. Checkmarks indicate which fields are injected into the memory block for each strategy. Macro is the arithmetic mean across the four benchmarks.}
  \label{tab:strategy}
  \centering
  \small
  \setlength{\tabcolsep}{7pt}
  \begin{tabular}{@{}lccccccccc@{}}
    \toprule
    Strategy & Q & A & E & Annot & BizB & F-E & F-M & MME & Macro \\
    \midrule
    A & \checkmark & & & & 54 & 67 & 53 & 38 & 53.00 \\
    B & \checkmark & \checkmark & & & 58 & 66 & 61 & 40 & 56.25 \\
    C & \checkmark & \checkmark & \checkmark & & 66 & 71 & 64 & 47 & 62.00 \\
    D & \checkmark & \checkmark & & \checkmark & 65 & 64 & 63 & 42 & 58.50 \\
    E & \checkmark & \checkmark & \checkmark & \checkmark & 69 & 81 & 67 & 53 & 67.50 \\
    \bottomrule
  \end{tabular}
\end{table*}

The strategy comparisons in Table~\ref{tab:strategy} decompose the contribution of each field. Adding gold answers to Q-only references, A to B, yields 3.25 macro points, confirming that reference answers carry a measurable signal. Adding tools and annotation without experience, B to D, yields 2.25 points, less than half the gain from experience alone.

Experience provides the dominant marginal improvement. Strategy C, which includes Q, A, and experience but omits annotation, exceeds strategy D, which includes Q, A, and annotation but omits experience, by 3.50 macro points. This gap isolates Findings and Cautions from the effect of annotation. The gain is concentrated on FinMME, from 42 to 47, and on FinMMR Easy, from 64 to 71, benchmarks on which sign convention and unit conversion errors recur systematically and are captured by distilled guard rules.

Annotation amplifies experience rather than adding independently to it. Strategy E, the full pipeline, exceeds C by 5.50 points, a margin larger than the D-to-C gap of 3.50. This asymmetry indicates that annotator directives are most effective when experience content is already rich: personalization prevents a guard rule about fiscal year 2022 from being misapplied to a query about fiscal year 2024, but generic Findings and Cautions must first be present to be personalized. The total gain from A to E is 14.50 macro points, from 53.00 to 67.50. Taken together, the five strategies show that tools, experience, and annotation make complementary contributions.

\begin{figure}[t]
  \centering
  \includegraphics[width=\linewidth]{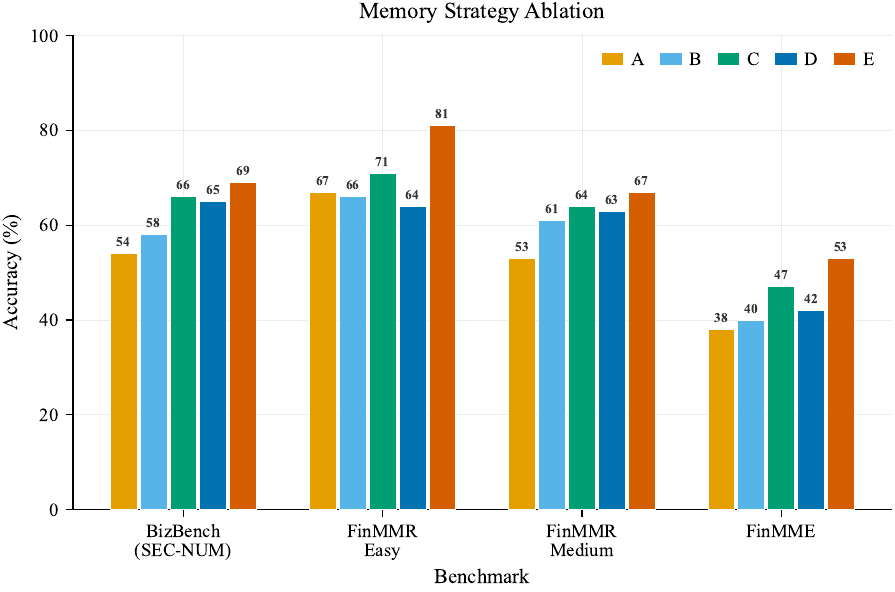}
  \caption{Memory strategy ablation on a shared test subsample. Checkmarks in Table~\ref{tab:strategy} indicate the injected fields for each strategy.}
  \label{fig:strategy}
\end{figure}

\section{Retrieval Stability Analysis}
\label{app:retrieval_stability}

\subsection{Experimental Design}
\label{sec:retrieval_exp_design}

We evaluate retrieval composition across 11 bank snapshots, from 0 to 1200 entries, by computing cosine similarity between all 11,379 test query embeddings and each bank entry at the operating threshold $\tau=0.65$. The snapshots are constructed as cumulative prefixes drawn uniformly across six benchmarks: entries from \path{finacumen/memory/v8_unified/meta.json} are partitioned by \texttt{source.dataset}, shuffled with seed 42, and the first $S/6$ entries per benchmark are concatenated for each target size $S \in \{0, 120, 240, \dots, 1200\}$. Each successive snapshot retains all entries from the previous size and appends 20 additional entries per benchmark, yielding a cumulative prefix of 120 entries per step. A small fraction of entries lack resolvable embeddings, yielding actual bank cardinalities slightly below the nominal target sizes. While we fix the primary operational threshold at $\tau=0.65$ for cross-benchmark analysis, we also sweep $\tau \in [0.50, 0.90]$ to examine how retrieval density and volatility scale under different matching strictness.

\subsection{Retrieval Metrics}
\label{sec:retrieval_metrics}

We report four metrics, where $R_n(q)$ denotes the set of entries retrieved for query $q$ at snapshot $n$ and $Q$ denotes the full query set of size 11{,}379.

\[
H_n = \frac{1}{|Q|}\sum_{q\in Q} \mathbf{1}[|R_n(q)| > 0]
\]

Hit rate $H_n$ measures the fraction of queries receiving at least one retrieved entry at bank size $n$.

\[
\bar{E}_n = \frac{1}{|Q|}\sum_{q\in Q} |R_n(q)|
\]

Mean entries per query $\bar{E}_n$ captures retrieval density.

For adjacent snapshots $n_1$ and $n_2$, the per-query Jaccard similarity is defined with an explicit case for two empty retrieval sets:

\[
J_q(n_1, n_2) =
\left\{
\begin{array}{l@{}l}
1, \quad \enspace\enspace\text{if } R_{n_1}(q) = R_{n_2}(q) = \emptyset,\\
\dfrac{|R_{n_1}(q) \cap R_{n_2}(q)|}
       {|R_{n_1}(q) \cup R_{n_2}(q)|}, \enspace\;\text{otherwise}.
\end{array}
\right.
\]

\[
J_{n_1,n_2} = \frac{1}{|Q|}\sum_{q\in Q} J_q(n_1, n_2)
\]

Adjacent-size Jaccard similarity $J_{n_1,n_2}$ quantifies compositional stability across snapshots. Assigning $J_q = 1$ when both retrieval sets are empty inflates Jaccard estimates for benchmarks with low hit rates, a phenomenon of trivial stability discussed in Section~\ref{sec:retrieval_stability_analysis}.

\[
C_{n_1,n_2} = \frac{1}{|Q|}\sum_{q\in Q} \mathbf{1}[R_{n_1}(q) \neq R_{n_2}(q)]
\]

Change fraction $C_{n_1,n_2}$ captures retrieval volatility as the strict proportion of queries whose retrieval set differs between snapshots.

\begin{figure}[t]
  \centering
  \includegraphics[width=\linewidth]{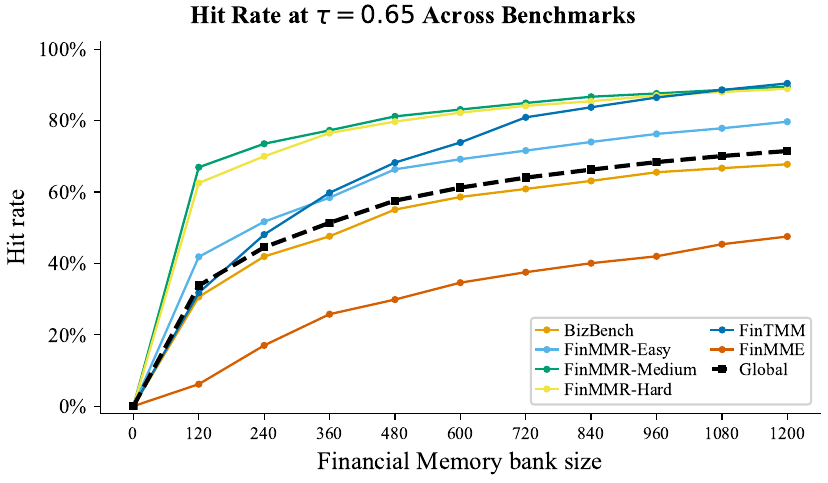}
  \caption{Hit rate as a function of \fm{} bank size at $\tau=0.65$, with a global aggregate curve. All benchmarks exhibit saturation beyond 600 entries; cross-dataset variation reflects divergent semantic affinity between benchmark query distributions and the bank's memorized patterns.}
  \label{fig:retrieval_hitrate}
\end{figure}

Per-dataset hit rate and Jaccard curves, together with their global aggregates, appear in Figures~\ref{fig:retrieval_hitrate} and~\ref{fig:retrieval_jaccard}. Table~\ref{tab:retrieval_state} summarizes the terminal-state metrics at bank size 1200 for $\tau=0.65$.

\begin{table}[t]
  \centering
  \caption{Terminal retrieval state at bank size $N=1200$ under $\tau=0.65$. $H_{1200}$ denotes hit rate, $\bar{E}_{1200}$ mean entries per query, and $J_{1080,1200}$ the Jaccard similarity at the final adjacent-size transition.}
  \label{tab:retrieval_state}
  \begin{tabular}{lccc}
    \toprule
    Benchmark & $H_{1200}$ & $\bar{E}_{1200}$ & $J_{1080,1200}$ \\
    \midrule
    BizBench        & 0.68 &  5.72 & 0.93 \\
    FinMMR-Easy     & 0.80 &  9.63 & 0.93 \\
    FinMMR-Medium   & 0.90 & 54.54 & 0.90 \\
    FinMMR-Hard     & 0.89 & 38.16 & 0.91 \\
    FinTMM          & 0.90 &  3.97 & 0.91 \\
    FinMME          & 0.48 &  1.15 & 0.95 \\
    \midrule
    Global          & 0.71 & 13.07 & 0.93 \\
    \bottomrule
  \end{tabular}
\end{table}

\subsection{Coverage and Retrieval Density}
\label{sec:retrieval_coverage}

Hit rate saturates beyond 600 entries across all benchmarks: FinTMM reaches 90.4\% at 1200 entries whereas FinMME reaches only 47.5\%. This gap reflects divergent semantic affinity between benchmark query distributions and the patterns memorized in the bank. Chart-oriented FinMME queries activate substantially fewer stored experiences than numerical reasoning queries, consistent with a bank composition that over-represents text- and table-centric failure cases. Mean entries per query confirms this pattern: at $\tau=0.65$, FinMMR-Medium and FinMMR-Hard retrieve 54.5 and 38.2 entries per query on average, while FinMME retrieves 1.1. FinTMM presents an intermediate profile with a hit rate of 90.4\% but a mean entry count of only 4.0, indicating that its queries resolve with sparser yet highly targeted retrieval sets.

\begin{figure}[t]
  \centering
  \includegraphics[width=\linewidth]{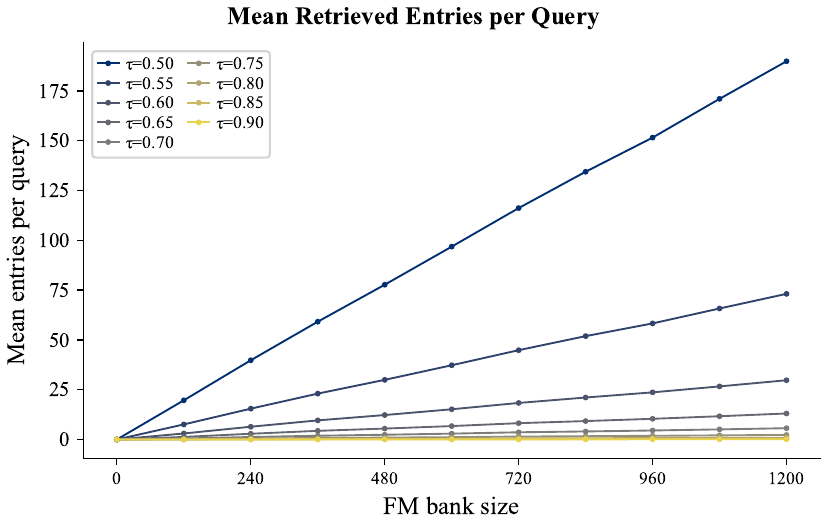}
  \caption{Mean number of retrieved entries per query aggregated across all 11,379 test queries. Each curve corresponds to one cosine threshold; the near-linear growth is predicted by $E[|R_n(q)|] = n \cdot p$.}
  \label{fig:retrieval_mean_entries}
\end{figure}

Under threshold-based retrieval, the expected number of retrieved entries follows $E[|R_n(q)|] = n \cdot p$, where $p = P(\text{sim}(q, e) \geq \tau)$ is independent of the current bank size. This produces the nearly linear growth of mean entries observed in Figure~\ref{fig:retrieval_mean_entries}. Because new entries are purely additive under threshold filtering, the retrieval set monotonically expands: $R_n(q) \subseteq R_{n+\Delta n}(q)$ holds for all queries.

\subsection{Retrieval Stability}
\label{sec:retrieval_stability_analysis}

\begin{figure}[t]
  \centering
  \includegraphics[width=\linewidth]{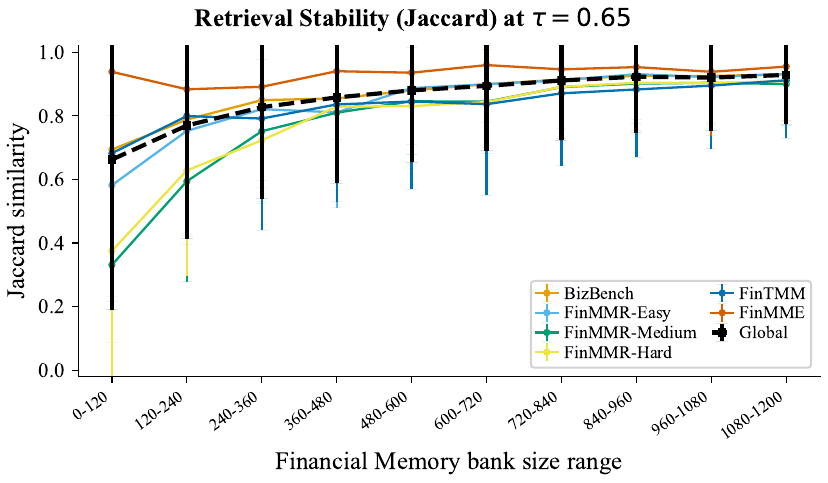}
  \caption{Adjacent-bank Jaccard similarity at $\tau=0.65$ with error bars indicating one standard deviation across queries. Jaccard similarity increases monotonically toward 1.0, consistent with $E[J] \approx n / (n + \Delta n)$; minor local fluctuations are attributable to sample variance in newly added entries.}
  \label{fig:retrieval_jaccard}
\end{figure}

Since $R_n(q) \subseteq R_{n+\Delta n}(q)$ holds under threshold retrieval, the expected Jaccard similarity between adjacent snapshots has the closed form
\[
E[J_{n, n+\Delta n}] \approx \frac{n \cdot p}{(n + \Delta n) \cdot p} = \frac{n}{n + \Delta n},
\]
which increases monotonically and approaches 1.0 as $n$ grows. This matches the dominant trend in Figure~\ref{fig:retrieval_jaccard}: all benchmarks exhibit a monotonic rise toward saturation, with the final transition values exceeding 0.90 for every dataset. Minor terminal fluctuations, such as the 0.004 drop observed in FinMMR-Medium, are consistent with sample variance: at specific transitions the empirical retrieval probability $\hat{p}$ of newly admitted entries deviates slightly from the population $p$, producing sub-thousandth-level oscillations that are not statistically significant.

The high Jaccard values of FinMME require careful interpretation. FinMME exhibits the lowest hit rate among all benchmarks, reaching only 47.5\% at 1200 entries. For the remaining 52.5\% of FinMME queries no entries are retrieved at either snapshot, and the convention $J_q(\emptyset, \emptyset) = 1$ applies. This trivial stability inflates the observed Jaccard toward 1.0 and partially accounts for FinMME's higher Jaccard curve. For benchmarks with sufficient retrieval coverage the Jaccard trajectory reflects genuine compositional convergence.

\begin{figure}[t]
  \centering
  \includegraphics[width=\linewidth]{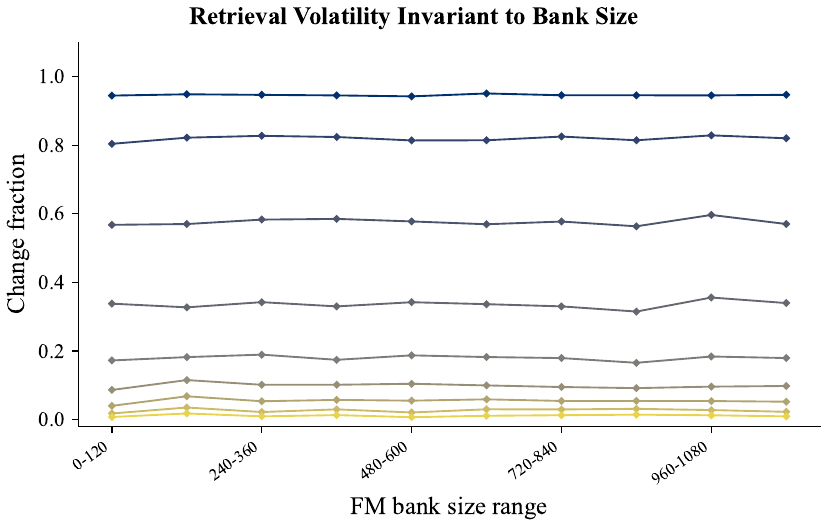}
  \caption{Change fraction across adjacent bank snapshots for nine cosine thresholds. The near-horizontal trajectory for each threshold confirms that retrieval volatility is invariant to bank size, as predicted by $P(\text{Change}) = 1 - (1-p)^{\Delta n}$.}
  \label{fig:retrieval_change_fraction}
\end{figure}

The change fraction $C_{n_1,n_2}$ measures the proportion of queries for which at least one newly admitted entry exceeds the similarity threshold. Under threshold retrieval, a query's retrieval set changes if and only if at least one entry among the $\Delta n = 120$ newly added entries satisfies the similarity condition. This probability is
\[
P(\text{Change}) = 1 - (1 - p)^{\Delta n},
\]
which depends solely on $\tau$ and the fixed step size $\Delta n$, not on the current bank size $n$. Empirically, the change fraction curves in Figure~\ref{fig:retrieval_change_fraction} are near-horizontal across all bank size transitions for every threshold value, confirming this invariance.

Taken together, these results establish that under threshold-based retrieval, retrieval stability is an inherent consequence of the retrieval mechanism rather than an emergent property of bank size. Jaccard similarity converges toward 1.0 as $n/(n+\Delta n)$ approaches unity, and change fraction remains constant at $1 - (1-p)^{\Delta n}$ independent of $n$. For practitioners, adjusting the similarity threshold $\tau$ offers a more direct lever for controlling retrieval churn than limiting bank size.

\section{Benchmark Examples}
\label{app:benchmark_examples}

\begin{figure*}[t]
  \centering
  \includegraphics[width=\textwidth]{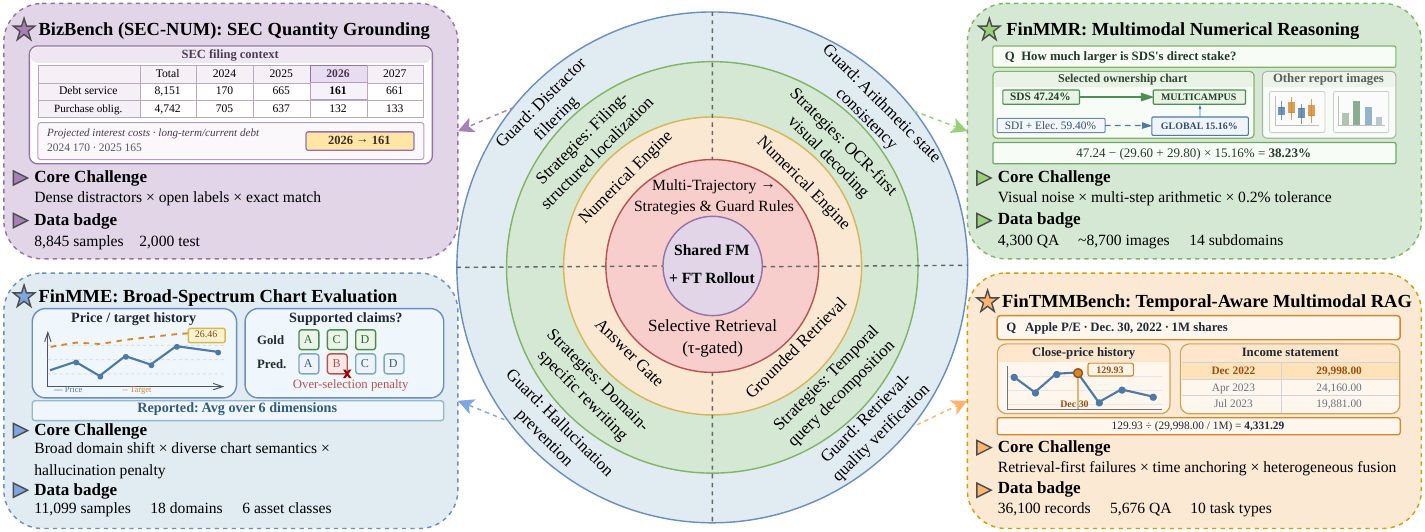}
  \caption{Example items from the four financial multimodal benchmarks evaluated in this work.}
  \label{fig:benchmark_examples}
\end{figure*}

\section{Experimental Details}
\label{app:impl}

This appendix documents the experimental protocol behind Table~\ref{tab:main} and the ablation analyses. The goal is to make the comparison reproducible while keeping the main text focused on the central findings.

\subsection{Evaluation Protocols}
\label{app:eval_protocols}
\label{app:finmme_metric}

We evaluate \fa{} on the official test split of each benchmark and follow the native metric defined by the corresponding benchmark paper. For BizBench SEC-NUM, predictions are normalized by removing currency markers, percentage signs, commas, and whitespace before exact-match comparison with a 1\% relative numerical tolerance. For FinMMR, we report accuracy on the Easy, Medium, and Hard splits using the official 0.2\% relative tolerance after numeric and unit normalization. For FinTMMBench, we use the benchmark's GPT-4o-mini judge to decide whether the prediction and the reference answer express the same financial fact. For FinMME, we report Avg as the arithmetic mean of Compre, FG, Reason, Single, Multi, and Cal. The reconstruction and consistency check are provided below.

All reported \fa{} results follow these native scoring rules. We do not introduce a unified post-hoc metric across datasets because the benchmarks differ substantially in answer form, ranging from exact quantities to temporally grounded retrieval answers and chart-based multiple-choice questions.

\begin{table*}[t]
  \centering
  \footnotesize
  \setlength{\tabcolsep}{3pt}
  \caption{Benchmark settings used in the main experiments. Each benchmark keeps its original split and native scoring rule.}
  \label{tab:dataset_protocols}
  \begin{tabular}{@{}
    >{\raggedright\arraybackslash}p{0.17\textwidth}
    >{\raggedright\arraybackslash}p{0.17\textwidth}
    >{\centering\arraybackslash}p{0.09\textwidth}
    >{\raggedright\arraybackslash}p{0.29\textwidth}
    >{\raggedright\arraybackslash}p{0.19\textwidth}
  @{}}
    \toprule
    Benchmark & Setting & Test size & Primary challenge & Reported metric \\
    \midrule
    BizBench SEC-NUM & SEC filings & 2000 & Numeric grounding under distractors & Accuracy, 1\% tolerance \\
    FinMMR & Easy / Medium / Hard & 1200 / 1200 / 1000 & Multi-image financial reasoning & Accuracy, 0.2\% tolerance \\
    FinTMMBench & Temporal multimodal QA & 1136 & Retrieval over tables, news, prices, and charts & GPT-4o-mini judge accuracy \\
    FinMME & Chart-intensive financial QA & 2220 & Chart understanding across financial domains & Avg \\
    \bottomrule
  \end{tabular}
\end{table*}

Following Table 3 of FinMME~\citep{luo2025finmme}, we compute Avg as the arithmetic mean of the six reported dimensions:
\[
\begin{aligned}
\mathrm{Avg}=\frac{1}{6}\bigl(&\mathrm{Compre}+\mathrm{FG}+\mathrm{Reason}\\
&+\mathrm{Single}+\mathrm{Multi}+\mathrm{Cal}\bigr).
\end{aligned}
\]
Applying the published dimension scores reproduces the reported Avg values within rounding.

\begin{table*}[t]
  \centering
  \caption{Consistency check for the FinMME Avg metric.}
  \label{tab:finmme_reconstruction}
  \begin{tabular}{lrrr}
    \toprule
    Model & Recomputed & Published & Difference \\
    \midrule
    Gemini 2.0 Flash  & 51.8500 & 51.85 & 0.000 \\
    Claude 3.5 Sonnet & 48.1950 & 48.20 & 0.005 \\
    GPT-4o            & 46.5583 & 46.56 & 0.002 \\
    Qwen2.5-VL-72B    & 52.5417 & 52.54 & 0.002 \\
    \bottomrule
  \end{tabular}
\end{table*}

The released annotations do not provide official per-sample domain labels. We therefore reconstruct the six-dimension partition, and the resulting cognitive split differs from Table 2 of FinMME by at most 0.02 percentage points.

\subsection{FinAcumen Configuration}
\label{app:finacumen_config}

The backbone for all \fa{} and Base rows is Qwen3-VL-8B-Instruct. The backbone parameters are frozen throughout; \fa{} changes only the inference-time context and tool execution path through \fm{} and \ft{}. Unless otherwise stated, decoding uses temperature 0 for final test evaluation, and the ReAct-style solver is capped at 16 tool-interaction steps. The tool suite is fixed across all \fa{} runs and consists of Python execution for deterministic arithmetic, financial data lookup, OCR-based visual structure extraction, and a termination gate that checks source traceability, unit consistency, and answer format.

For each source instance, memory construction samples $K=8$ independent trajectories before scoring and consolidation. The memory bank is built before final testing from training/validation trajectories and is kept read-only during evaluation. Test questions never write new entries into the bank. Each memory entry stores the source question, gold answer, distilled findings from successful trajectories, and cautions extracted from failed trajectories. At inference time, the query embedding is formed from the question and the first 600 characters of the context, then compared against pre-computed memory-entry embeddings. Entries below the cosine threshold $\tau=0.65$ are discarded, duplicate source items are removed, and at most $k_{\max}=5$ retrieved entries are rendered into the memory prefix for the main results in Table~\ref{tab:main}. The Base row disables both \fm{} and \ft{}, while ablations keep the backbone, decoding configuration, tool interfaces, and benchmark scoring fixed. All other retrieval and tool parameters follow the architecture described in Section~\ref{sec:method}.

\subsection{Baseline Attribution}
\label{app:baselines}

The non-\fa{} baseline numbers in Table~\ref{tab:main} are taken from the original benchmark papers rather than re-run in our environment: BizBench results from \citet{bizbench}, FinMMR results from \citet{tang2025finmmr}, FinTMMBench results from \citet{zhu2025towards}, and FinMME results from \citet{luo2025finmme}. This choice follows the standard leaderboard-comparison protocol for these benchmarks and avoids conflating our framework contribution with differences in closed-source API versions, prompting details, or evaluation access. When a benchmark paper reports a model family rather than a single shared variant across all columns, Table~\ref{tab:main} groups the results by family and the per-cell model variant is specified below.

\begin{table*}[t]
  \centering
  \scriptsize
  \setlength{\tabcolsep}{3pt}
  \caption{Per-cell model variants for grouped General LLM baselines in Table~\ref{tab:main}. All numbers are copied from the corresponding benchmark papers. A dash indicates that the source benchmark did not report the model.}
  \label{tab:baseline_variants}
  \begin{tabular}{@{}
    >{\raggedright\arraybackslash}p{0.13\textwidth}
    >{\centering\arraybackslash}p{0.12\textwidth}
    >{\centering\arraybackslash}p{0.13\textwidth}
    >{\centering\arraybackslash}p{0.13\textwidth}
    >{\centering\arraybackslash}p{0.13\textwidth}
    >{\centering\arraybackslash}p{0.13\textwidth}
    >{\centering\arraybackslash}p{0.13\textwidth}
  @{}}
    \toprule
    Model family & BizBench & FinMMR-Easy & FinMMR-Medium & FinMMR-Hard & FinTMMBench & FinMME \\
    \midrule
    GPT & GPT-4 & GPT-4o & GPT-4o & GPT-4o & GPT-4o-mini & GPT-4o \\
    Claude Sonnet & Claude 3.7 Sonnet & Claude 3.7 Sonnet & Claude 3.7 Sonnet & Claude 3.7 Sonnet & -- & Claude 3.5 Sonnet \\
    Llama & Llama 2 70B & Llama 4 Maverick 17B & Llama 4 Maverick 17B & Llama 4 Maverick 17B & Llama 3.2 11B & -- \\
    \bottomrule
  \end{tabular}
\end{table*}

For Qwen2.5-VL-72B, Gemini~2.0 Flash, and the financial-domain baselines, we preserve the exact variants and scores reported by the benchmark sources whenever available. Missing cells in Table~\ref{tab:main} mean that the benchmark paper did not report that model under the official setting, not that the model failed evaluation.

\section{Repeated-Run Uncertainty}
\label{app:uncertainty}

We report measurement uncertainty for Base and \fa{} from five independent runs under identical conditions. Each cell contains the mean, empirical standard deviation, and 95\% confidence interval. The interval is computed as
\[
\bar{x} \pm t_{0.975,4}\frac{s}{\sqrt{5}},
\qquad
t_{0.975,4}=2.776.
\]

\begin{table*}[t]
  \centering
  \caption{Repeated-run uncertainty for Base and \fa{} across five independent runs. Each cell reports the mean, standard deviation, and 95\% confidence interval.}
  \label{tab:uncertainty}
  \begin{tabular}{llrr}
    \toprule
    Benchmark & System & Mean $\pm$ std & 95\% CI \\
    \midrule
    BizBench      & Base  & $27.67 \pm 1.00$ & $[26.43, 28.91]$ \\
                  & \fa{} & $68.65 \pm 2.45$ & $[65.61, 71.69]$ \\
    FinMMR Easy   & Base  & $59.17 \pm 1.64$ & $[57.13, 61.21]$ \\
                  & \fa{} & $81.67 \pm 1.14$ & $[80.25, 83.09]$ \\
    FinMMR Medium & Base  & $45.75 \pm 1.30$ & $[44.14, 47.36]$ \\
                  & \fa{} & $67.00 \pm 0.71$ & $[66.12, 67.88]$ \\
    FinMMR Hard   & Base  & $31.40 \pm 1.00$ & $[30.16, 32.64]$ \\
                  & \fa{} & $46.10 \pm 2.86$ & $[42.55, 49.65]$ \\
    FinTMMBench   & Base  & $16.20 \pm 0.55$ & $[15.52, 16.88]$ \\
                  & \fa{} & $19.98 \pm 0.89$ & $[18.88, 21.08]$ \\
    FinMME        & Base  & $27.84 \pm 0.84$ & $[26.80, 28.88]$ \\
                  & \fa{} & $51.08 \pm 2.24$ & $[48.30, 53.86]$ \\
    \bottomrule
  \end{tabular}
\end{table*}

The reported intervals are separated in each evaluated setting. \fa{} has lower empirical variance on FinMMR Easy and FinMMR Medium and higher empirical variance on the remaining settings.

\section{Experience-Memory Baselines}
\label{app:memory_baselines}

We compare \fa{} with Reflexion~\citep{shinn2023reflexion} and ExpeL~\citep{zhao2024expel}. The three methods use Qwen3-VL-8B-Instruct as the backbone, share the same tool environment and training data for experience construction, and freeze their memory banks before evaluation while retaining their respective experience representations. This protocol focuses the comparison on how experience is represented and used.

\begin{table}[t]
  \centering
  \caption{Comparison with experience-memory baselines under a shared backbone and tool environment.}
  \label{tab:memory_baselines}
  \begin{tabular}{llr}
    \toprule
    Setting & Method & Accuracy \\
    \midrule
    \multirow{3}{*}{FinMMR E}
      & Reflexion & 77.17 \\
      & ExpeL     & 77.17 \\
      & \fa{}     & 81.67 \\
    \midrule
    \multirow{3}{*}{FinMMR M}
      & Reflexion & 63.58 \\
      & ExpeL     & 60.58 \\
      & \fa{}     & 67.00 \\
    \midrule
    \multirow{3}{*}{FinTMM}
      & Reflexion & 14.52 \\
      & ExpeL     & 16.73 \\
      & \fa{}     & 19.98 \\
    \midrule
    \multirow{3}{*}{FinMME}
      & Reflexion & 42.97 \\
      & ExpeL     & 45.77 \\
      & \fa{}     & 51.08 \\
    \bottomrule
  \end{tabular}
\end{table}

The four comparisons favor \fa{} under this protocol. This pattern may reflect the separation of successful strategies and failure patterns into Findings and Cautions, together with experience activation conditioned on sufficient relevance. Appendix~\ref{app:strategy} provides complementary component evidence. With Q, A, Annot, and the tool configuration unchanged, adding the Experience field increases macro accuracy from 58.50 to 67.50.

\section{Cross-Backbone and Scale Generalization}
\label{app:cross_backbone}

\subsection{Additional Backbones}

We evaluate three additional backbones on BizBench under the configuration used for Table~\ref{tab:main}.

\begin{table}[H]
  \centering
  \caption{BizBench results with additional backbones.}
  \label{tab:cross_backbone}
  \begin{tabular}{lrr}
    \toprule
    Base model & Base & \fa{} \\
    \midrule
    GLM-4.6V              & 30.40 & 75.58 \\
    GPT-4o-mini           & 23.60 & 59.60 \\
    Gemini-2.5-Flash-Lite & 32.60 & 48.40 \\
    \bottomrule
  \end{tabular}
\end{table}

The observed differences are positive for the three evaluated backbones, providing evidence that the performance gain is not confined to Qwen3-VL-8B-Instruct.

\subsection{Larger Backbone}

We further evaluate Qwen3-VL-235B-A22B-Instruct on the 2000-item BizBench SEC-NUM test set. The prompt, tool interfaces, and decoding configuration remain fixed across the three settings.

\begin{table}[H]
  \centering
  \setlength{\tabcolsep}{2pt}
  \caption{Component results with Qwen3-VL-235B-A22B-Instruct on BizBench SEC-NUM.}
  \label{tab:qwen235b}
  \begin{tabular}{lrrr}
    \toprule
    Setting & Accuracy & $\Delta$ Base & $\Delta$ FT only \\
    \midrule
    Base     & 57.45 & --    & --   \\
    FT only  & 69.30 & 11.85 & --   \\
    \fa{}    & 75.10 & 17.65 & 5.80 \\
    \bottomrule
  \end{tabular}
\end{table}

FT only improves accuracy by 11.85 points over Base, and Financial Memory adds 5.80 points over FT only. The result is consistent with the complementary pattern observed in the 8B ablation.

\section{Cost Analysis}
\label{app:cost}

\subsection{Offline Construction Cost}

\fa{} samples $K=8$ independent trajectories per source instance. Based on measurements from six source instances, we estimate the offline cost of constructing the approximately 1800-entry memory bank in Table~\ref{tab:construction_cost}. The sequential time was measured on an Intel Core i7-12700H processor. The bank is constructed and frozen before evaluation, so trajectory sampling, scoring, and consolidation are not repeated during inference.

\begin{table}[H]
  \centering
  \caption{Estimated offline memory construction cost for the approximately 1800-entry bank.}
  \label{tab:construction_cost}
  \begin{tabular}{lr}
    \toprule
    Measure & Estimate \\
    \midrule
    API calls       & 64,200 \\
    Tokens          & 244.43 million \\
    Cost            & CNY 142.97 \\
    Sequential time & 78.12 hours \\
    Storage         & 8.19 MiB \\
    \bottomrule
  \end{tabular}
\end{table}

\subsection{Inference Cost}

We sample one sixth of the test instances from each of the six evaluation settings and conduct paired runs with and without the bank. Table~\ref{tab:inference_cost} reports the average inference cost per question.

\begin{table}[H]
  \centering
  \caption{Average inference cost per question in paired runs.}
  \label{tab:inference_cost}
  \begin{tabular}{lrr}
    \toprule
    Measure & With bank & Without bank \\
    \midrule
    Input tokens  & 40,462.02 & 19,072.23 \\
    Output tokens &  2,061.11 &    971.28 \\
    Total tokens  & 42,523.13 & 20,043.51 \\
    \midrule
    Input cost  & CNY 0.0202 & CNY 0.0095 \\
    Output cost & CNY 0.0041 & CNY 0.0019 \\
    Total cost  & CNY 0.0244 & CNY 0.0115 \\
    \bottomrule
  \end{tabular}
\end{table}

Memory activation increases the average total token count from 20,043.51 to 42,523.13 and the average cost from CNY 0.0115 to CNY 0.0244 per question. The approximately 1800-entry bank is constructed once and shared across queries. Selective retrieval bypasses annotation and memory injection when no experience satisfies the retrieval condition.

\section{Prompt Templates}
\label{app:prompts}

\begin{figure}[t]
  \centering
  \includegraphics[width=\columnwidth]{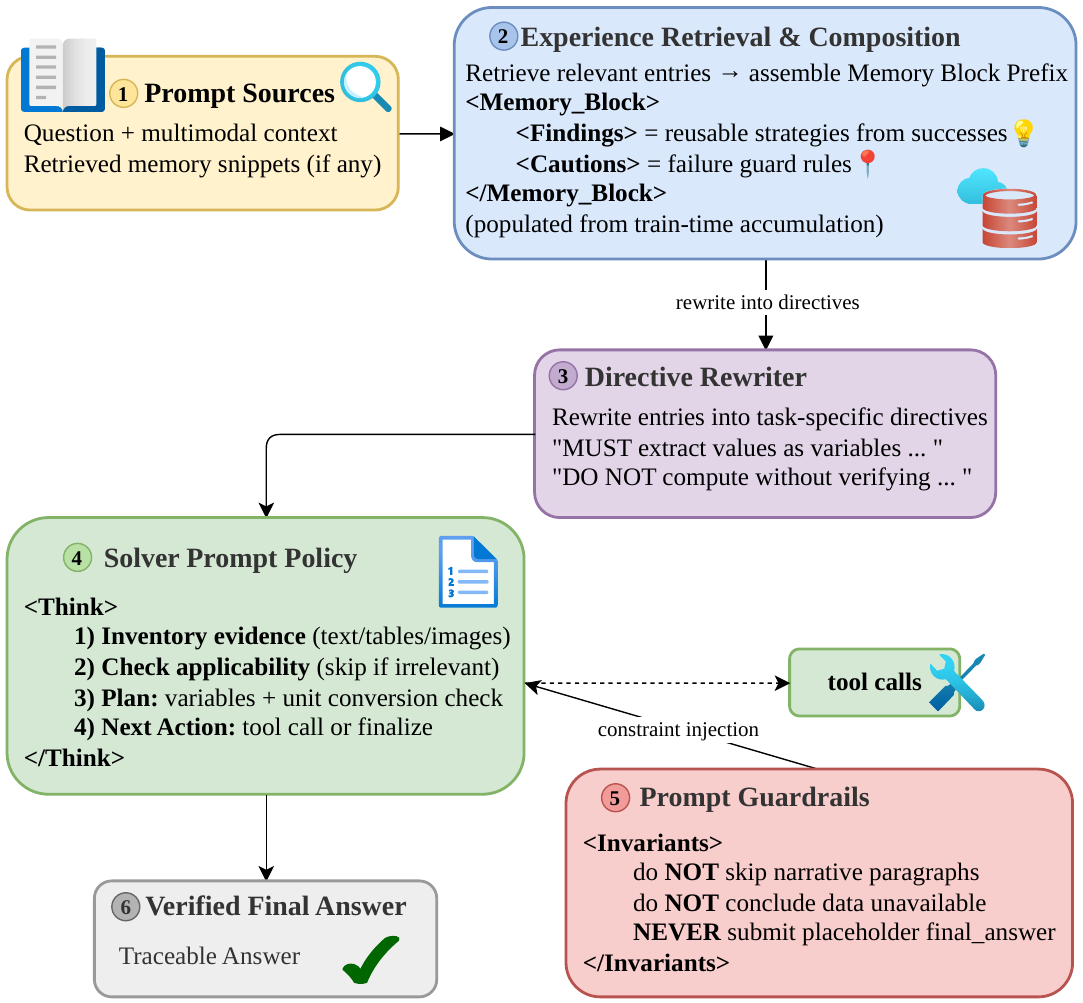}
  \caption{Prompt orchestration in \fa{}.}
  \label{fig:prompt}
\end{figure}

Figure~\ref{fig:prompt} provides a visual overview of the prompt orchestration in \fa{}. Each prompt excerpt is labeled by its corresponding agent role,
described in \S\ref{sec:method:agent}.

\begin{tcolorbox}[
  title=\textbf{Solver agent (FinAcumen System Prompt, excerpted)},
  colback=gray!8,
  breakable,
  fonttitle=\small\bfseries,
]
\begin{lstlisting}[style=promptstyle]
<Memory_Handling>
The input may contain a <Memory_Block> before the <Problem> block.
Each <Entry> has:
- <Question>: The original problem that produced this experience.
  Use it to judge whether your current problem is in a similar scenario
  (industry, data source, problem shape).
- <Answer>: The correct answer to that original problem.
  Use as a format reference --- not to copy the value, but to understand
  what kind of output your problem expects (number, text, percentage, etc.).
- <Experience>: Distilled reusable rules organized as
  <Findings> and <Cautions>:
    <Findings>: Patterns that succeeded --- follow these strategies.
    <Cautions>: Guard rules from failures --- apply if the condition triggers.

The block ends with an <OptOut> --- use it only when an entry truly
does not fit.
</Memory_Handling>

<Think_Steps>
<Think>
### 1. Input Inventory (MANDATORY --- do NOT skip to computation)
Output a complete inventory before any tool call.
[NARRATIVE] Every standalone number with its unit and context, one per line.
[TABLES] ALL row labels with ALL column values; list each cell.
[IMAGES] ALL fields from OCR/vision; flag ambiguous readings.
[TERM_MATCH] Map question terms to inventory entries.
  If a term has NO direct match, list candidate matches explicitly.

If any section is incomplete, you have not finished reading.
DO NOT proceed to Strategy until the inventory covers ALL of Context.

### 2. Experience Applicability (if Memory_Block present)
{step2_text}

### 3. Problem Understanding
- What quantity? Unit? Precision? Check Instruction.
- Resolve TERM_MATCH warnings.
- Output TYPE: ENTITY NAME, NUMERIC VALUE, DESCRIPTIVE SENTENCE,
  VERDICT, or NEWS SENTIMENT. Match answer format to expected type:
  * ENTITY -> terminate(final_answer="Company Name")
  * NUMERIC -> terminate(final_answer="VALUE UNIT")
  * DESCRIPTIVE -> terminate(final_answer="Direction: description with numbers")
  * VERDICT -> terminate(final_answer="Yes/No" or "true/false")
  * NEWS -> terminate(final_answer="Yes" or "No")

### 4. Strategy
Write ALL values as python variables before computing
(e.g., rev_2022 = 125.4).
Check unit conversion (millions<>billions, %<>decimal).

### 5. Next Action
Tool call or terminate.
</Think>
After </Think>, emit the tool call. No &lt;Answer&gt; block.
</Think_Steps>

<Invariants>
- NEVER skip narrative paragraphs to jump to tables.
  Context is one document --- prose AND tables must both be read.
- "data not available" is never true. If you think data is missing,
  re-read from the beginning --- you missed a section.
- NEVER submit placeholder text as final_answer: avoid
  'I will terminate', 'data not available', 'insufficient data',
  empty strings.
- final_answer MUST contain the actual answer value, not a meta-status.
</Invariants>
\end{lstlisting}
\end{tcolorbox}

\begin{tcolorbox}[
  title=\textbf{Summary agent---Experience synthesis (excerpted)},
  colback=gray!8,
  breakable,
  fonttitle=\small\bfseries,
]
\begin{lstlisting}[style=promptstyle]
<findings_guide>
Write generalized reusable strategies. Strip ALL specific values:
numbers, years, entity names.
Format: "To <goal>, use/do <method>."
Trigger: what information format triggers this? What question type?
Method: what to do step by step? Which tool? Which formula?
Each finding is one sentence, self-contained and executable.

<cautions_guide>
Write generalized guard rules. Strip ALL specific values.
Format: "When <detectable condition>, <corrective action>."
Condition must be detectable from question/context text
(not hindsight).
Action must be concrete: what to check, how to check, what to do.

<Output>
Return ONLY:
<Think>
analysis: [problem-specific derivation]
-> From above, generalize findings:
-> From above, generalize cautions:
-> Verify: does each item have trace support?
   Is the condition detectable?
</Think>
<Answer>
{"findings": ["To ..., use/do ..."],
 "cautions":  ["When ..., ..."],
 "analysis":  "<<REPLACE_WITH_YOUR_PROBLEM_DERIVATION>>"}
</Answer>
\end{lstlisting}
\end{tcolorbox}

\begin{tcolorbox}[
  title=\textbf{Annotator agent---Directive annotation (excerpted)},
  colback=gray!8,
  breakable,
  fonttitle=\small\bfseries,
]
\begin{lstlisting}[style=promptstyle]
<Task>
For each entry write 1-3 directive sentences using MUST / DO NOT.
Extract the entry's methodology (extraction pattern, computation logic,
verification step) and adapt it to the current problem.
NEVER write an empty annotation.
Set "useful": false ONLY when the task TYPE is fundamentally different
(chart reading vs. table lookup vs. narrative extraction).
Otherwise, useful MUST be true.

<Generalization_Rule>
Map the memory's METHODOLOGY, not its literal entity names.
Memory: specific years/entity names -> current problem: relevant equivalents.
If you cannot map entities, the entry is NOT useful.

<Output>
[{"i": 0, "useful": true,
  "annotation": "MUST extract values from Context table as python
   variables before computing. DO NOT compute without verifying which
   table row matches the question."}, ...]
\end{lstlisting}
\end{tcolorbox}

\begin{tcolorbox}[
  title=\textbf{Memory\_Block XML injection format},
  colback=gray!8,
  breakable,
  fonttitle=\small\bfseries,
]
\begin{lstlisting}[style=promptstyle]
<Memory_Block>
  <Field_Guide>
  Each Entry is a past experience:
    - Question: the original problem (use to judge scenario similarity)
    - Answer: the correct answer (format reference only, never copy value)
    - Experience: distilled reusable rules organized as:
      - Findings: patterns that succeeded -> follow these strategies
      - Cautions: guard rules from failures -> apply if condition triggers
  </Field_Guide>
  <Entry>
    <Question>original problem text</Question>
    <Answer>correct answer</Answer>
    <Experience>
      <Findings>reusable strategies from successes</Findings>
      <Cautions>guard rules from failures</Cautions>
    </Experience>
  </Entry>
  <OptOut>If any entry above does NOT actually fit this problem,
    ignore it.</OptOut>
</Memory_Block>
\end{lstlisting}
\end{tcolorbox}

\end{document}